\documentclass[sensors,article,accept,moreauthors,pdftex,10pt,a4paper]{Definitions/mdpi}

\firstpage{1}
\pubvolume{18}
\issuenum{8}
\articlenumber{2485}
\pubyear{2018}
\copyrightyear{2018}
\history{Received: 15 June 2018; Accepted: 26 July 2018; Published: 31 July 2018}

\updates{yes} 
\usepackage{bm}

\usepackage{booktabs}
\usepackage{multirow}
\usepackage{soul}
\usepackage{microtype}

\DeclareMathOperator*{\argmin}{arg\,min}

\Title{Attributes’ Importance for Zero-Shot Pose-Classification Based on Wearable Sensors}

\Author{{Hiroki Ohashi} $^{1,}$*, Mohammad Al-Naser $^{2}$, Sheraz Ahmed $^{2}$\href{https://orcid.org/0000-0002-4239-6520}{\orcidicon}, Katsuyuki Nakamura $^{1}$, Takuto~Sato~$^{1}$ and Andreas Dengel $^{2}$}

\AuthorNames{Hiroki Ohashi, Mohammad Al-Naser, Sheraz Ahmed, Katsuyuki Nakamura, Takuto Sato, and Andreas Dengel}

\address{%
$^{1}$ \quad Research \& Development Group, Hitachi, Ltd.,  {Tokyo 185-8601, Japan};  katsuyuki.nakamura.xv@hitachi.com~(K.N.);  takuto.sato.hn@hitachi.com (T.S.)\\
$^{2}$ \quad German Research Center for Artificial Intelligence (DFKI), { 67663 Kaiserslautern}, Germnay; Mohammad\_Osamh\_Adel.Al-Naser@dfki.de~(M.A.-N.); Sheraz.Ahmed@dfki.de (S.A.);  Andreas.Dengel@dfki.de (A.D.)}

\corres{Correspondence: hiroki.ohashi.uo@hitachi.com; Tel.: +81-42-323-1111}

\abstract{
This paper presents a simple yet effective method for improving the performance of zero-shot learning (ZSL).
ZSL classifies instances of unseen classes, from which no training data is available, by utilizing the attributes of the classes.
Conventional ZSL methods have equally dealt with all the available attributes, but this sometimes causes misclassification.
This is because an~attribute that is effective for classifying instances of one class is not always effective for another class.
In this case, a metric of classifying the latter class can be undesirably influenced by the irrelevant attribute.
This paper solves this problem by taking the importance of each attribute for each class into account when calculating the metric.
In addition to the proposal of this new method, this~paper also contributes by providing a dataset for pose classification based on wearable sensors, named~HDPoseDS.
It contains 22 classes of poses performed by 10 subjects with 31 IMU sensors across full body.
To the best of our knowledge,  it is the richest wearable-sensor dataset especially in terms of sensor density, and thus it is suitable for studying zero-shot pose/action recognition.
The~presented method was evaluated on HDPoseDS and outperformed relative improvement of 5.9\% in comparison to the best baseline method.
}


\begin{document}
\section{Introduction}
\label{sec:intro}
Human-action recognition (HAR) has wide range of applications such as life log, healthcare, video surveillance, and worker assistance.
The recent advances in deep neural networks (DNN) have drastically enhanced the performance of HAR both in terms of recognition accuracy and coverage of the recognized actions
\cite{Herath2017,wang2017deep}.
DNN-based methods, however, sometimes face difficulty in practical deployment; a system user sometimes wants to change or add target actions to be recognized, but it is not so trivial for DNN-based methods to do so since they require large amount of training data of the new target actions.

Zero-shot learning (ZSL) has a great potential to overcome this difficulty of dependence on training data when recognizing a new target class \cite{larochelle2008zero,Frome2013,Fu2018}.
Whereas in normal supervised-learning setting, the set of classes contained in test data is exactly the same as that in training data, it is not the case in ZSL; test data includes ``unseen'' classes, of which instances are not contained in training data.
In other words, if $\mathcal{Y}_{train}$ is a set of class labels in training data and $\mathcal{Y}_{test}$ is that in test data, then~$\mathcal{Y}_{train} = \mathcal{Y}_{test}$ in normal supervised-learning framework, while $\mathcal{Y}_{train} \neq \mathcal{Y}_{test}$ in ZSL framework.
(more specifically, $\mathcal{Y}_{train} \cap \mathcal{Y}_{test} = \phi$ in some cases, and $\mathcal{Y}_{train} \subset \mathcal{Y}_{test}$ in other cases).
Unseen classes are classified using {\it attribute} together with a description of the class based on the attributes, which~ is usually given on the basis of external knowledge.
Most typically it is manually given by humans~\cite{lampert2009learning, lampert2014attribute}.
The {\it attribute} represents a semantic property of the class.
A classifier to judge the presence of the attribute (or the probability of the presence) is learnt using training data.
For example, the attribute of ``striped'' can be learnt using the data of ``striped shirts'', while the attribute of ``four-legged'' can be learnt using the data of ``lion''.
Then an unseen class ``zebra'' can be recognized, without any training data of zebra itself, by using these attribute classifiers as well as the description that zebras are striped and four-legged.

The idea of ZSL has been applied also to human action recognition \cite{Liu2011,cheng2013towards,Xu2016,li2016recognizing,qin2017zero}.
Indeed they successfully demonstrated a capability of recognizing unseen actions, but the attributes used in these studies are relatively task-specific and not so fundamental as to be able to recognize wider variety of human actions.
The potential of recognizing truly wide variety of actions becomes substantially bigger if more fundamental and general set of attributes are utilized.
To this end, we believe the status of each human-body joint is appropriate attribute since any kinds of human action can be  represented using the set of each body joint's status.

There are sophisticated vision-based methods such as \cite{iqbal2017posetrack,chen2017adversarial,guler2018densepose} for estimating the status of body joint, but the problem of occlusion is essentially inevitable for those approach.
Moreover, they are not suitable for the applications in which the target person moves around beyond the range of camera view.
Thus we utilize wearable sensors, which are free from occlusion problem, to estimate the statuses of all the major human body joints.
We aim at developing a method that flexibly recognizes wide range of human actions with ZSL.
This study especially focuses on the classification of static actions, or poses, as the first step toward that goal {{(Some of the} poses are sometimes referred to as ``action'' in prior works, but we use the term ``pose'' in this study)}.

The biggest challenge in zero-shot pose recognition is the intra-class variation of the poses.
The~difficulty of intra-class variation in general action recognition was discussed in \cite{Liu2011}.
For~example, when ``folding arm'', one may clench his/her fists while another may not.
The authors introduced a method to deal with the intra-class variation by regarding attributes as latent variables.
However, it was for normal supervised learning and their implementation in ZSL scenario was naive nearest-neighbor-based method that does not address this problem.
The intra-class variation becomes an even severe problem in ZSL especially when fine-grained attributes like each body joint's status are utilized.
This is because the value of all the attributes should be specified in ZSL even though some of the attribute actually may take arbitrary values.
It is difficult to uniquely define the status of hands for ``folding arm'', but the attribute ``hands'' cannot be omitted because it is necessary for recognizing other poses such as ``pointing''.
Conventional ZSL methods have dealt with all the attributes equally even though some of them are actually not important for some of the classes.
This sometimes causes misclassification because a metric (e.g., likelihood, distance) to represent that a given sample belongs to a class can be undesirably influenced by irrelevant attributes.
This paper solves the problem by taking the importance of each attribute for each class into account when calculating the metric.

The effectiveness of the method is demonstrated on a human pose dataset collected by us that is named Hitachi-DFKI pose dataset, or HDPoseDS in short.
HDPoseDS contains 22 classes of poses performed by 10 subjects with 31 inertial measurement unit (IMU) sensors across full body.
To the best of our knowledge, this is the richest dataset especially in terms of sensor density for human pose classification based on wearable sensors.
Due to its sensor density, it gives us a chance of extracting fundamental set of attributes for human poses, namely the status of body joints.
Therefore, it is the first dataset suitable for studying wearable-based zero-shot learning in which wide variety of full-body poses are involved.
We make this dataset publicly available to encourage the community for further research in this direction.
It is available at \url{http://projects.dfki.uni-kl.de/zsl/data/}.

The main contribution of this study is two folds.
(1) We present a simple yet effective method to enhance the performance of ZSL by taking the importance of each attribute for each class into account. We experimentally show the effectiveness of our method in comparison to baseline methods.
(2) We provide HDPoseDS, a rich dataset for human pose classification suitable especially for studying wearable-based zero-shot learning.
In addition to these major contributions, we also present a practical design for estimating the status of each body joint; while conventional ZSL methods formulate attribute-detection problem as 2-class classification (whether the attribute is present or not), we~estimate it under the scheme of either multiclass classification or regression depending upon the characteristics of each body joint.

\section{Related Work}
\label{sec:related work}
We review three types of prior works in this section, namely ZSL, wearable-based action and pose recognition, and wearable-based zero-shot action and pose recognition.

\subsection{Zero-Shot Learning}
The idea of ZSL was firstly presented in \cite{larochelle2008zero} followed by \cite{lampert2009learning} and \cite{palatucci2009zero}.
The major input sources have been images and videos, but there have been some studies based on wearable sensors as reviewed later in this section.
The most fundamental framework established in the early days is as follows.
Firstly a~function $f: \mathcal{X} \mapsto \mathcal{A}$ is learnt using labeled training data, where $\mathcal{X}$ denotes an input (feature) space, and $\mathcal{A}$ denotes an attribute space.
The definition of unseen classes is given manually, and it represents a vector in the attribute space $\mathcal{A}$.
Then a function $g: \mathcal{A} \mapsto \mathcal{Y}$ is learnt using the vectors in $\mathcal{A}$ and their labels.
Here $\mathcal{Y}$ denotes a label space.
When test data are given, their labels are estimated by applying the learnt functions $f$ and $g$ subsequently.
In early days, SVM was frequently used for learning $f$, and~it's replaced by DNN-based method these days.
One of the most common methods for learning $g$ has been nearest neighbors \cite{palatucci2009zero,Liu2011,cheng2013nuactiv,xu2015semantic}.
This study also uses a nearest-neighbor-based method.

Extensive efforts have been made to improve ZSL methods from various viewpoints.
Socher~et~al.~\cite{socher2013zero} invented a method that does not need manual definition of unseen classes by utilizing natural language corpora (word2vec).
Jayaraman and Grauman~\cite{jayaraman2014zero} took the unreliability of attribute estimation into account by a random-forest based method.
Semantic representations were effectively enriched by using synonyms in \cite{alexiou2016exploring}, and by using textual descriptions as well as relevant still images in \cite{wang2017alternative}.
Qin~et~al.~\cite{qin2016beyond} extended the semantic attributes to latent attributes in order to obtain more discriminative representation as well as more balanced attributes.
Tong~et~al.~\cite{tong2017adversarial} were the first to introduce generative adversarial network (GAN)~\cite{goodfellow2014generative} in ZSL.
A problem of domain shift that is common in ZSL was effectively dealt with in \cite{Xu2016} and \cite{qin2017zero}.
Liu~et~al.~\cite{liu2018cross} studied cross-modal ZSL between tactile data and visual data.
Our idea of incorporating each attribute's importance for each class was inspired by \cite{jayaraman2014zero} as their idea of incorporating attributes' unreliability is similar in terms of dealing with the characteristics of attributes.

\subsection{Wearable-Based Action and Pose Recognition}
As reviewed in \cite{lara2013survey,bulling2014tutorial,mukhopadhyay2015wearable}, the mainstream of action recognition methods before DNN-based methods become popular has consisted of two-stage approach; firstly they apply a sliding window to time-series data and extract time domain features such as mean and standard deviation as well as frequency domain features such as FFT coefficients, and secondly apply various machine-learning method such as hidden Markov model (HMM) \cite{pmlr-v48-guan16}, support vector machine (SVM) \cite{bulling2012multimodal}, conditional random field (CRF) \cite{adams2016hierarchical}, and an ensemble method  \cite{zheng2013physical}.

In recent years, DNN-based approaches have become increasingly popular as they showed overwhelming results \cite{wang2017deep}.
In \cite{yang2015deep,jiang2015human,ronao2015deep}, they introduced a way to employ convolutional neural networks (CNN) to automatically extract efficient features from time-series data.
Ord{\'o}{\~n}ez~et~al.~\cite{ordonez2016deep} proposed a method to more explicitly deal with the temporal dependencies of the human actions by utilizing long short-term memory (LSTM).
Hammerla~et~al.~\cite{hammerla2016deep} also introduced a LSTM-based method and gave the performance comparison among DNN, CNN, and LSTM as well as the influence of the network parameters in each method.
Following the findings from these researches, we also utilize CNN for estimating the status of each body joint.
The details of the implementation will be given in the following section.

\subsection{Wearable-Based Zero-Shot Action and Pose Recognition}
One of the earliest attempts to apply the idea of ZSL to human activity recognition based on wearable sensors is \cite{cheng2013nuactiv} and their subsequent work \cite{cheng2013towards}.
They firstly used nearest-neighbor-based approach to recognize activities using attributes, and later enhanced the method to incorporate temporal dependency by using CRF.
Wang~et~al.~\cite{wang2017zero} proposed a nonlinear compatibility based method, where they first project the features extracted from sensor readings to a hidden space by a nonlinear function, and then calculate the compatibility score based on the features in the hidden space and prototypes in semantic space.
Al-Naser~et~al.~\cite{naser2017} introduced a ZSL model for recognizing complex activities by using simpler actions and surrounding objects as attributes.

These prior works successfully showed a great potential of realizing zero-shot action recognition based on wearable sensors.
However, on one hand the attributes used in those studies are neither fine-grained nor fundamental enough so as to represent truly wide variety of human actions or poses.
On the other hand, the methods used in those studies do not take the attributes' importance into account, which matters more especially when using fine-grained attributes to represent diverse poses.

\section{Dataset: HDPoseDS}
\vspace{-6pt}

\subsection{Sensor}
\begin{figure}[H]
\begin{center}
\includegraphics[width=0.9\columnwidth]{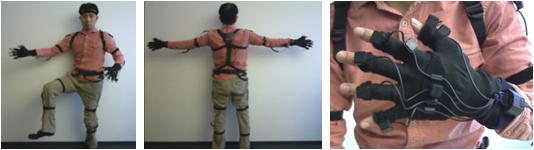}
\vspace{-5pt}
\caption{Sensor displacement in Perception Neuron.}
\label{fig:pn}
\end{center}
\end{figure}

Our goal is to use all the major human-body joints as attributes to represent full-body poses.
Thus, a very dense sensor set across full-body is required.
Perception Neuron from Noitom Ltd. is ideal for this {purpose} (\url{https://neuronmocap.com/}).
It has 31 IMU sensors across full body; 1 on head, 2~on shoulders, 2 on upper arms, 2 on lower arms, 2 on hands, 14 on fingers, 1 on spine, 1 on hip, 2~on upper legs, 2~on lower legs, 2 on feet (Figure~\ref{fig:pn}).
Each IMU is composed of a 3-axis accelerometer, 3-axis gyroscope and 3-axis magnetometer.
We use 10 dimensional data from each IMU including 3~acceleration data, 3 gyro data, and 4 quaternion data.
This rich set of sensors are especially helpful in applications where detailed full-body pose-recognition is desired.
For example, in workers' training, novice workers can learn to avoid undesirable poses that can cause safety or quality issues with the help of a pose recognition system.

\subsection{Target Poses}
\begin{figure}[H]
\begin{center}
\includegraphics[width=0.8\linewidth]{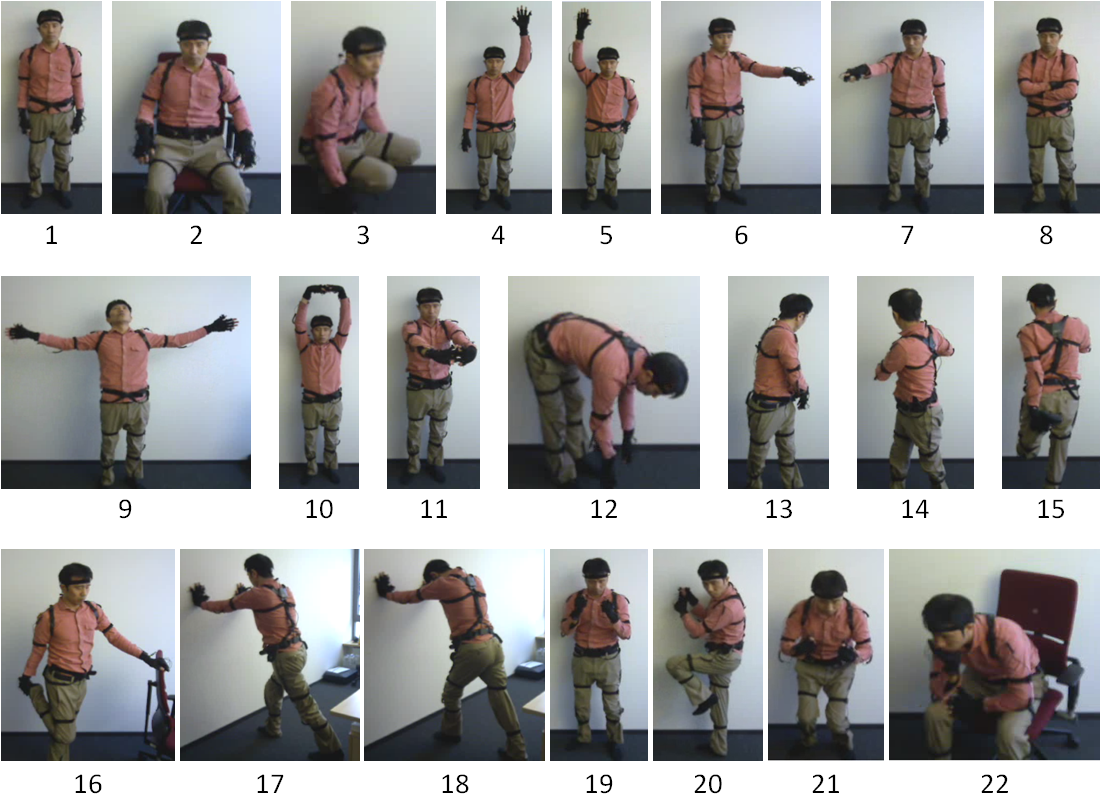}
\vspace{-8pt}
\caption{Poses in HDPoseDS. The numbers under the figures correspond to the numbers in Table~\ref{tab:val}.}
\label{fig:poses}
\end{center}
\end{figure}

\textls[-15]{In order to test the generalization capability of zero-shot models, we constructed a human pose dataset named HDPoseDS using Perception Neuron.
We newly built the dataset because existing wearable-sensor datasets are not collected with so densely-attached sensors as to be used for extracting fundamental set of attributes for human poses, namely the status of each body joint.
We defined 22 poses such that various body parts are involved and thus the generalization capability of the developed method in zero-shot scenario can be appropriately tested (see Figure~\ref{fig:poses} for appearance and Table~\ref{tab:val}~for~names)}.

\vspace{-12pt}

\begin{table}[H]
\centering
\caption{Intra-class variation observed in data collection. Poses are sometimes represented in slightly different manner depending on the subjects. L and R in the pose names means Left and Right.}
\label{tab:val}

\begin{tabular}{llll}
\toprule
\textbf{ID} & \textbf{Pose}                    & \textbf{Variation}                                                                   & \textbf{Involved Body Joint} \\ \midrule
1  & Standing               & no big variation                     & - \\ \midrule
2  & Sitting                  & hands on a table, on knees, or straight down                    & elbows, hands \\ \midrule
3  & Squatting              & hands hold on to sth, on knees, or straight down               & elbows, hands \\ \midrule
4, 5 & Raising arm (L, R)           & a hand on hip, or straight down                                      & elbow, hand \\ \midrule
6, 7 & Pointing (L, R)               & a hand on hip, or straight down                                      & elbow, hand \\ \midrule
8  & Folding arm           & \begin{tabular}[c]{@{}l@{}}wrist curled or straight, \\ hands clenched or normal\end{tabular}                              & \begin{tabular}[c]{@{}l@{}}wrist, \\ hands\end{tabular} \\ \midrule
9  & Deep breathing      & head up or front                                                          & head \\ \midrule
10 & Stretching up        & head up or front                                                          & head \\ \midrule
11 & Stretching forward & waist straight or half-bent                                             & waist \\ \midrule
12 & Waist bending       & no big variation                     & - \\ \midrule
13, 14 & Waist twisting (L, R)        & \begin{tabular}[c]{@{}l@{}}head left (right) or front, \\ arms down or left (right)\end{tabular} & \begin{tabular}[c]{@{}l@{}}head,\\  shoulders, elbows\end{tabular} \\ \midrule
15, 16 & Heel to back (L, R)        & \begin{tabular}[c]{@{}l@{}}a hand hold on to sth, straight down, \\ or stretch horizontally\end{tabular} & shoulder, elbow, hand \\ \midrule
17, 18 & Stretching calf (L, R)      & head front or down                                                       & head \\ \midrule
19 & Boxing                 & head front or down                                                       & head \\ \midrule
20 & Baseball hitting     & head left or front                                                          & head \\ \midrule
21 & Skiing                  & head front or down                                                       & head \\ \midrule
22 & Thinking               & \begin{tabular}[c]{@{}l@{}}head front or down,\\   wrist reverse curled or normal, \\ hand clenched or normal\end{tabular} & \begin{tabular}[c]{@{}l@{}}head, \\ wrist, \\ hands\end{tabular}    \\ \bottomrule
\end{tabular}
\end{table}

We had 10 subjects, and each subject performed all the 22 poses for about 30 s.
All of the 10~subjects were males, but from 4 different countries.
The body heights of the subjects ranged from 160~cm to 185~cm.
The ages were from 23 years old to 37 years old.
To the best of our knowledge, this~is the richest dataset especially in terms of sensor density (31 IMU across full body).
Therefore, it~is the first dataset suitable for studying wearable-based zero-shot learning in which wide variety of full-body poses are involved.
We make this dataset publicly available to encourage the community for further research in this direction.
It is available at \url{http://projects.dfki.uni-kl.de/zsl/data/}.

During the data collection, only brief explanation about each pose was given, and thus we observed some intra-class variation in the dataset as summarized in Table~\ref{tab:val}.

\section{Proposed Method}
\label{sec:proposed method}
Following the standard scheme of ZSL, our approach also consists of two stages; attribute estimation based on sensor readings, and class label estimation using estimated attributes. We explain the two stages one by one in detail.
In this section, we first explain the sensor to be used in our study, then describe the two stages in detail.

\subsection{Attribute Estimation}
\label{sec:ae}
We use 14 major human-body joints as attributes to represent various poses as summarized in the first column of Table~\ref{tab:poses}.
Unlike conventional ZSL methods, where 2-class classification is always used (whether an attribute is present or not), we use either multiclass classification or regression depending upon the characteristics of each body joint as shown in Table~\ref{tab:poses}.
For the joints that have only one degree of freedom like knees (or at least whose major movement is restricted to one dimension), it is more suitable and beneficial to use regression to estimate the status.
This allows us to represent intermediate status of the joint by just specifying an intermediate value, which enables to describe detailed status of the joint, rather than just ``straight'' and ``curl'',  to represent more complicated poses in the future.
For~the joints that have more than 2 degrees of freedom like head, we use multiclass classification.
It is indeed possible to replace this by 2-class classification on each status, but it's more natural to formulate this as a multiclass classification problem since each status are mutually exclusive (e.g., if head is ``up'', then it cannot be ``down'' at the same time).

\begin{figure}[H]
\begin{center}
\includegraphics[width=0.95\linewidth]{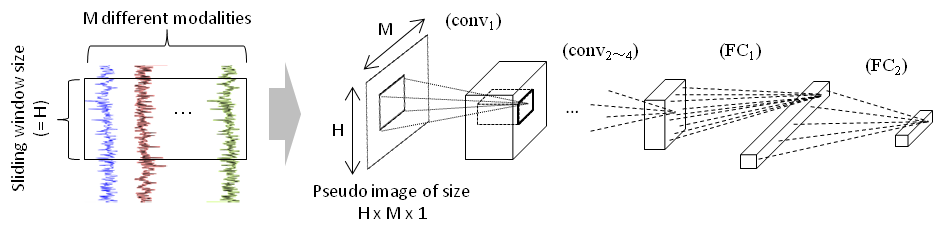}
\caption{Architecture of time-series CNN for basic pose recognition (attribute estimation).}
\label{fig:cnn}
\end{center}
\end{figure}

We use CNN to deal with multivariate time-series data.
Previous studies \cite{yang2015deep,ordonez2016deep,hammerla2016deep} first dealt with different modalities individually by applying convolution only on temporal direction (a kernel size of CNN is $k\times1$), and integrated the output from all the modality in fully connected layers that appear right before the classification layer.
However, as shown in Figure~\ref{fig:cnn}, we integrate the readings from different sensor modalities in the earlier stage using CNN  (a kernel size of CNN is $k_1\times k_2$) since~we experimentally found that it gives better performance.
We construct one network per one joint, resulted in 14 networks to estimate the status of all the joints.

\begin{table}[H]
\centering
\caption{The 14 body joints used as attributes and the types and values to represent their status.
Note~that each joint has left part and right part except head and waist.}
\begin{tabular}{lll}\toprule
\textbf{Joint}    & \textbf{Type} & \textbf{Value}\\ \midrule
head    & classification & up, down, left, right, front \\
shoulder & classification & up, down, left, right, front \\
elbow & regression & 0 (straight)--1 (bend) \\
wrist & regression & 0 (reverse curl)--1 (curl)\\
hand & classification & normal, grasp, pointing \\
waist & classification & straight, bend, twist-L, twist-R\\
hip joint & regression & 0 (straight)--1 (bend) \\
knee & regression & 0 (straight)--1 (bend) \\ \bottomrule
\end{tabular}
\label{tab:poses}
\end{table}

The sliding window size in this study is 60, which corresponds to 1 s.
The number of modality (referred to as $M$ in Figure~\ref{fig:cnn}) is $s \times 10$, where $s$ is the number of used IMU for each joint.
We use 4 convolution layers with 25, 20, 15, and 10 channels.
One fully connected layer with 100 nodes is inserted before the final classification or regression layer.
The activation function is leaky ReLU throughout the network but the regression layer has sigmoid activation to squash the values to $[0, 1]$.
We use cross entropy loss for multiclass classification, and mean absolute loss for regression.
They are optimized using MomentumSGD with momentum value of 0.9.
Batch normalization and drop out is used for regularization.
The kernel size in convolution layers are $3\times10$ for hands and $3\times3$ for all the rest.
We use the wider kernel for hands because the number of IMU used for classifying hands' status is significantly larger than that for other joints.


\subsection{Pose Classification with Attributes' Importance}
\label{sec:ai}\vspace{-6pt}
\subsubsection{Naive Formulation}
We use nearest-neighbor-based method for zero-shot pose classification.
The input to pose classification is the output from attribute estimation explained in Section~\ref{sec:ae}.
The output dimension from the networks for each joint is the number of classes if the joint's status is estimated by multiclass classification, and 1 if it is by regression, resulting in a 33 dimensional vector in total.
Let \mbox{$\bm{a^{(n)}}=\{a^{(n)}_1, \cdots, a^{(n)}_D\}$} be an attribute vector of $n$'th sample ($D=33$ in this study).
Please note that $\forall a^{(n)}_d \in [0, 1]$ since we squash the values by softmax and sigmoid function for multiclass classification and regression, respectively.
For ZSL, we need extra information for estimating pose labels using vectors in attribute space.
Following some of the previous works \cite{lampert2009learning,lampert2014attribute,cheng2013towards,cheng2013nuactiv}, we simply use a manually defined table for this as shown in Table~\ref{tab:def} and convert them to corresponding 33-dimensional vectors (one-of-K representation is used for multiclass classification part).

For normal nearest-neighbor-based method, a given test data $\bm{x}$ is first converted to an attribute vector $\bm{a}$ using the learnt attribute estimation networks, and then the distance between $\bm{a}$ and the $i$th training data $\bm{v}^{(i)}$ is calculated as follows.
\begin{equation}
\label{eq:d1}
d(\bm{a}, \bm{v}^{(i)}) = (\sum_d^{D}{|a_d - v^{(i)}_d|^p})^{1/p},
\end{equation}
where $D$ is the dimension of an attribute space, and $x_d$ denotes the $d$'th element of vector $\bm{x}$.
For seen classes, $\bm{v^{(i)}}\in\mathbb{R}^{D}$ is a training data mapped to the attribute space using the learnt attribute estimation networks, while for unseen classes it is a vector created based on the pose definition table (Table~\ref{tab:def}).
$p$~is usually 1 or 2.
Then the given data $\bm{x}$ is classified to the following class.
\begin{equation}
C(\argmin_{i} {d(\bm{a}, \bm{v}^{(i)})}), 
\end{equation}
where $C(i)$ gives the class to which $i$th training data belong.

\tabcolsep = 2pt
\begin{table}[H]
\centering
\caption{Definition of the poses in HDPoseDS. (L) denotes left part and (R) denotes right part. For~joints, He: Head, S: Shoulder, E: Elbow, Wr: Wrist, Ha: Hand, Wa: Waist, HJ: Hip Joint, K: Knee. For joint status, F: Front, U: Up, D: Down, L: Left, R: Right, N: Normal, G: Grasp, P: Pointing, S: Straight, B:~Bend, TwL (R): Twist to Left (Right).}
\label{tab:def}
\small
\begin{tabular}{llcccccccccccccc}
\toprule
& \textbf{Pose\textbackslash Joint}  & \textbf{He} & \textbf{S(L)} & \textbf{S(R)} & \textbf{E(L)} & \textbf{E(R)} & \textbf{Wr(L)} & \textbf{Wr(R)} & \textbf{Ha(L)} & \textbf{Ha(R)} & \textbf{Wa}  & \textbf{HJ(L)} & \textbf{HJ(R)} & \textbf{K(L)} & \textbf{K(R)} \\ \midrule
1  & Standing           & F & D    & D    & 0    & 0    & 0.5   & 0.5   & N    & N    & S   & 0     & 0     & 0    & 0    \\
2  & Sitting            & F & D    & D    & 0    & 0    & 0.5   & 0.5   & N    & N    & S   & 0.5   & 0.5   & 0.5  & 0.5  \\
3  & Squatting          & F & D    & D    & 0    & 0    & 0.5   & 0.5   & N    & N    & S   & 1     & 1     & 1    & 1    \\
4  & Raising arm (L)       & F & U    & D    & 0    & 0    & 0.5   & 0.5   & N    & N    & S   & 0     & 0     & 0    & 0    \\
5  & Raising arm (R)       & F & D    & U    & 0    & 0    & 0.5   & 0.5   & N    & N    & S   & 0     & 0     & 0    & 0    \\
6  & Pointing (L)               & F & F    & D    & 0    & 0    & 0.5   & 0.5   & P    & N    & S   & 0     & 0     & 0    & 0    \\
7  & Pointing (R)               & F & D    & F    & 0    & 0    & 0.5   & 0.5   & N    & P    & S   & 0     & 0     & 0    & 0    \\
8  & Folding arm        & F & D    & D    & 0.5  & 0.5  & 0.5   & 0.5   & N    & N    & S   & 0     & 0     & 0    & 0    \\
9  & Deep breathing     & F & L    & R    & 0    & 0    & 0.5   & 0.5   & N    & N    & S   & 0     & 0     & 0    & 0    \\
10 & Stretching up      & F & U    & U    & 0    & 0    & 0     & 0     & N    & N    & S   & 0     & 0     & 0    & 0    \\
11 & Stretching forward & F & F    & F    & 0    & 0    & 0     & 0     & N    & N    & S   & 0     & 0     & 0    & 0    \\
12 & Waist bending      & F & D    & D    & 0    & 0    & 0.5   & 0.5   & N    & N    & B   & 0     & 0     & 0    & 0    \\
13 & Waist twisting (L)  & L & L    & L    & 0    & 0.5  & 0.5   & 0.5   & N    & N    & TwL & 0     & 0     & 0    & 0    \\
14 & Waist twisting (R)  & R & R    & R    & 0.5  & 0    & 0.5   & 0.5   & N    & N    & TwR & 0     & 0     & 0    & 0    \\
15 & Heel to back (L)    & F & D    & D    & 0    & 0    & 0.5   & 0.5   & G    & N    & S   & 0     & 0     & 1    & 0    \\
16 & Heel to back (R)    & F & D    & D    & 0    & 0    & 0.5   & 0.5   & N    & G    & S   & 0     & 0     & 0    & 1    \\
17 & Stretching calf (L) & F & F    & F    & 0    & 0    & 0     & 0     & N    & N    & S   & 0     & 0.3   & 0    & 0.3  \\
18 & Stretching calf (R) & F & F    & F    & 0    & 0    & 0     & 0     & N    & N    & S   & 0.3   & 0     & 0.3  & 0    \\
19 & Boxing             & F & D    & D    & 1    & 1    & 0.5   & 0.5   & G    & G    & S   & 0     & 0     & 0    & 0    \\
20 & Baseball hitting   & L & D    & D    & 0.5  & 0.5  & 0.5   & 0.5   & G    & G    & S   & 0.5   & 0     & 0.5  & 0    \\
21 & Skiing             & F & D    & D    & 0.5  & 0.5  & 0.5   & 0.5   & G    & G    & S   & 0.3   & 0.3   & 0.3  & 0.3  \\
22 & Thinking           & F & D    & D    & 0.5  & 1    & 0.5   & 0     & N    & N    & S   & 1     & 1     & 0.5  & 0.5  \\ \bottomrule
\end{tabular}
\end{table}
\tabcolsep = 6pt

\subsubsection{Incorporating Attributes' Importance}
\label{subsubsec:ai}
As summarized in Table~\ref{tab:val}, sometimes there is intra-class variation in the poses.
In normal supervised learning setting, this intra-class variation can be naturally learnt as training data cover various instances of the given class.
However, it is not possible to do so in ZSL since there is no training data other than a definition table of the unseen classes.
In this situation, it is not appropriate to equally deal with all the attribute because not all the attributes are equally important for classifying a particular class.
For example, for ``squatting'' class, the status of hip joints and knees are important, and the values are expected to be always 1 (bent).
On the other hand, the status of elbows are not important since it is still ``squatting'' regardless of the status of elbows;
the values may be 1 (bent in order to hold on to something or to put hands on knees) or 0 (straight down to the floor).
In other words, the~attribute values of elbows do not matter to tell whether given test data belong to squatting class or not.
Please note that we cannot simply omit the attribute ``elbow'' because it is indeed necessary for other poses such as ``folding arm''.
Therefore, we need to design a new distance metric so that we can incorporate {\it each attributes' importance for each class}.

We formulate this as follows.
\begin{eqnarray}
\label{eq:d2}
d(\bm{a}, \bm{v}^{(i)}) =
\quad \frac{1}{W^{(i)}_{ai}}(\sum_d^{D}{w_{ai}(d, i) w_{rc}(d) |a_d - v^{(i)}_d|^p})^{1/p} + \lambda\frac{1}{W^{(i)}_{ai}},
\end{eqnarray}
\vspace{-12pt}
\begin{eqnarray}
\label{eq:ai}
W^{(i)}_{ai} = \sum_{j}w'_{ai}(j, i),
\end{eqnarray}
\vspace{-12pt}
\begin{eqnarray}
w_{rc}(d) =  
\begin{cases}
1 & \text{{\it if d'th attribute is given by regression}} \\
0.5 & \text{{\it otherwise}},
\end{cases}
\end{eqnarray}
where $w'_{ai}(j, i)$ denotes the importance of joint $j$ for the class $C(i)$.
It is given manually in this study as shown in Table~\ref{tab:ai}.
It may be indeed an extra work to manually define the attribute importance, but~actually it does not require too much extra time because anyway the attribute table (Table~\ref{tab:def}) should be manually defined as it was the case in many previous works.
In addition, it is neither difficult because it is natural to assume that the person, or the system user, who defines the attribute table (Table~\ref{tab:def}) has enough knowledge not only on the definition of the target poses but also on which attribute (body-joint status) is important for each pose.
It may be also possible to infer the attributes' importance either from training data or external resources (e.g., word embedding) rather than manually defining them, but it lies beyond the scope of this study at this moment.

We use binary values for attributes' importance for simplicity, but it is easy to extend it to continuous numbers.
$w_{ai}(d, i)$ is the importance of attribute $d$ for the class $C(i)$ and the value of it is copied from $w'_{ai}(j, i)$, where $d$'th attribute comes from joint $j$ .
Please note that it depends not only on $d$ but also on $C(i)$.
$w_{rc}(d)$ is 0.5 if the $d$'th attribute is calculated using multiclass classification because the total distance with regards to the joint $j$ from which the $d$'th attribute comes from $(\sum_{d \in joint j}{|a_d - v^{(i)}_d|})$ ranges from 0 to 2, whereas the distance with regard to the joint whose status is estimated using regression ranges from 0 to 1.
$W_{ai}$ can be interpreted as the total number of ``valid'' joints to be used for classification of class $C(i)$.
Therefore, the first term on the right-hand side in Equation~(\ref{eq:d2}) can be interpreted as the average distance between $\bm{a}$ and $\bm{v}^{(i)}$ over ``valid'' attribute that comes from ``valid'' joints.
The second term is introduced to penalize the class that uses too few attributes.
If~test data have the same distance to two classes that have different numbers of important attributes, this~term encourages to classify the data to the class which has larger number of important attributes, which~indicates more detailed definition of the pose.
We use $\lambda = 0.1$ and $p = 1$ in this study.
Note~that the increase of the computational cost compared to the naive formulation (Equation~(\ref{eq:d1})) is trivial because we just multiply constant numbers when calculating the distances.
\tabcolsep = 2pt
\begin{table}[H]
\centering
\caption{Attributes' importance.}
\label{tab:ai}
\small
\begin{tabular}{llcccccccccccccc}
\toprule
& \textbf{Pose}               & \textbf{He} & \textbf{S(L)} & \textbf{S(R)} & \textbf{E(L)} & \textbf{E(R)} & \textbf{Wr(L)} & \textbf{Wr(R)} & \textbf{Ha(L)} & \textbf{Ha(R)} & \textbf{Wa} & \textbf{HJ(L)} & \textbf{HJ(R)} & \textbf{K(L)} & \textbf{K(R)} \\ \midrule
1  & Standing           & 1 & 1    & 1    & 1    & 1    & 1     & 1     & 1    & 1    & 1  & 1     & 1     & 1    & 1    \\
2  & Sitting            & 1 & 1    & 1    & 0    & 0    & 0     & 0     & 0    & 0    & 1  & 1     & 1     & 1    & 1    \\
3  & Squatting          & 1 & 1    & 1    & 0    & 0    & 0     & 0     & 0    & 0    & 1  & 1     & 1     & 1    & 1    \\
4  & Raising arm (L)     & 1 & 1    & 1    & 1    & 0    & 1     & 0     & 1    & 0    & 1  & 1     & 1     & 1    & 1    \\
5  & Raising arm (R)     & 1 & 1    & 1    & 0    & 1    & 0     & 1     & 0    & 1    & 1  & 1     & 1     & 1    & 1    \\
6  & Pointing (L)        & 1 & 1    & 1    & 1    & 0    & 1     & 0     & 1    & 0    & 1  & 1     & 1     & 1    & 1    \\
7  & Pointing (R)        & 1 & 1    & 1    & 0    & 1    & 0     & 1     & 0    & 1    & 1  & 1     & 1     & 1    & 1    \\
8  & Folding arm        & 1 & 1    & 1    & 1    & 1    & 0     & 0     & 0    & 0    & 1  & 1     & 1     & 1    & 1    \\
9  & Deep breathing     & 0 & 1    & 1    & 1    & 1    & 1     & 1     & 1    & 1    & 1  & 1     & 1     & 1    & 1    \\
10 & Stretching up      & 0 & 1    & 1    & 1    & 1    & 1     & 1     & 1    & 1    & 1  & 1     & 1     & 1    & 1    \\
11 & Stretching forward & 1 & 1    & 1    & 1    & 1    & 1     & 1     & 1    & 1    & 0  & 1     & 1     & 1    & 1    \\
12 & Waist bending      & 0 & 0    & 0    & 0    & 0    & 0     & 0     & 0    & 0    & 1  & 1     & 1     & 1    & 1    \\
13 & Waist twisting (L)  & 1 & 0    & 0    & 0    & 0    & 0     & 0     & 0    & 0    & 1  & 1     & 1     & 1    & 1    \\
14 & Waist twisting (R)  & 1 & 0    & 0    & 0    & 0    & 0     & 0     & 0    & 0    & 1  & 1     & 1     & 1    & 1    \\
15 & Heel to back (L)    & 1 & 1    & 0    & 1    & 0    & 0     & 0     & 0    & 0    & 1  & 1     & 1     & 1    & 1    \\
16 & Heel to back (R)    & 1 & 0    & 1    & 0    & 1    & 0     & 0     & 0    & 0    & 1  & 1     & 1     & 1    & 1    \\
17 & Stretching calf (L) & 0 & 1    & 1    & 1    & 1    & 1     & 1     & 1    & 1    & 1  & 1     & 1     & 1    & 1    \\
18 & Stretching calf (R) & 0 & 1    & 1    & 1    & 1    & 1     & 1     & 1    & 1    & 1  & 1     & 1     & 1    & 1    \\
19 & Boxing             & 0 & 1    & 1    & 1    & 1    & 1     & 1     & 1    & 1    & 1  & 1     & 1     & 1    & 1    \\
20 & Baseball hitting   & 0 & 0    & 0    & 1    & 1    & 0     & 0     & 1    & 1    & 1  & 1     & 1     & 1    & 1    \\
21 & Skiing             & 0 & 1    & 1    & 1    & 1    & 1     & 1     & 1    & 1    & 0  & 1     & 1     & 1    & 1    \\
22 & Thinking           & 0 & 1    & 1    & 0    & 1    & 0     & 0     & 0    & 0    & 0  & 1     & 1     & 1    & 1    \\ \bottomrule
\end{tabular}
\end{table}
\tabcolsep = 6pt 

\section{Experiment}
\label{sec:experiment}
\vspace{-6pt}

\subsection{Evaluation Scheme}
We use HDPoseDS for evaluation.
The evaluation procedure is as follows.
\begin{enumerate}[leftmargin=7.6mm,labelsep=3mm]
\renewcommand{\labelenumi}{(\arabic{enumi}).}
\item	All the input data are converted to attribute vectors using the neural networks explained in Section~\ref{sec:ae}.
The sliding window size is 60, which corresponds to 1 s, and it's shifted by 30 (0.5 s).
This ends up with roughly 590 ($=(30/0.5-1) \times 10$) attribute vectors per pose since HDPoseDS contains data from 10 subjects and each subject performed roughly 30 s for each pose.
\item	For each class {\it c}, we construct a set of training data by combining the data from all the other classes than {\it c} and the pose definition of {\it c} based on attributes (Table~\ref{tab:def}).
We use class {\it c}'s data as test data.
\item	The labels of the test data are estimated using the method explained in Section~\ref{sec:ai}.
\item  We repeat this for all the 22 classes.
\item  We calculate the F-measure for each class based on the precision and recall rate.
\end{enumerate}

Please note that we do not assume that the possible output classes are only unseen (test) classes; we~assume that the seen (training) classes are also potential output classes during testing.
Since we do not use instances of seen (training) classes in testing, this evaluation scheme is not exactly the same as the generalized zero-shot learning (G-ZSL)~\cite{xian2017zero,kumar2018generalized}, in which the instances of seen classes are also used in testing.
It is, however, more similar to G-ZSL than normal ZSL in a sense that the target classes in testing include not only unseen (test) classes but also seen (training) classes.

In addition to this, we also investigate how the proposed method works in few-shot learning scenario, where only a small number of training data are available.
The evaluation procedure is the same as the ZSL case except the step (2);
instead of including the attribute definition of {\it c} in the training data, we include $k$ samples from class {\it c}'s data in $k$-shot learning scenario, and all the other data of class {\it c} are used for testing.
To choose the $k$ samples, firstly we randomly permutate class {\it c}'s data.
Then we use the $l$'th $(l=1,2, ..., \lfloor N_c / k \rfloor)$ $k$ samples for training and the remaining $(N_c - k)$ samples for testing, where $N_c$ is the number of class {\it c}'s data.
Then we proceed to step 3
The estimation result for class {\it c} is averaged, and then we proceed to step (4).

The performance of the proposed method is compared with three baseline methods.
The first one is one of the most frequently used method in ZSL studies, which is called ``direct attribute prediction (DAP)'' introduced in \cite{lampert2014attribute}.
Please note that we did not compare with indirect attribute prediction (IAP) that is also introduced in \cite{lampert2014attribute}.
This is because, as the authors of \cite{lampert2014attribute} stated, IAP is not appropriate for the case where training classes are also potential output class during testing.
The~other two~baseline methods are nearest-neighbor-based, which is also common in ZSL studies.
The~proposed method is also based on a nearest-neighbor method.
The first nearest-neighbor-based baseline is a naive nearest-neighbor-based method, in which the distance between samples are calculated using Equation~(\ref{eq:d1}) with weights $w_{rc}(d)$.
The second nearest-neighbor-based baseline is the one that uses random attributes' importance.
We~randomly generate either 0 or 1 for each $w'_{ai}(j, i)$ in Equation~(\ref{eq:ai}).
For this baseline method, we~test 1000 times using different random weights and report the average F-measure of the 1000 tests.
For~both of the proposed and the baseline methods, we use a~prototype representation (mean vector) of each class introduced in \cite{snell2017prototypical}, rather than all the training data themselves, in order to deal with the severe imbalance of number of training samples.
We tested all the method using a single desktop PC with Intel® Core™ i7-8700K CPU and NVIDIA GeForce GTX 1080 GPU.

\subsection{Results and Discussion}
The result of ZSL is summarized in Table~\ref{tab:reszsl}.
{The details are given} in Appendix \ref{sec:appendix}. 
As shown in the table, our proposed method outperformed all the baseline methods in average F-measure.
In addition, the proposed method could run at about 20 Hz, which is near real-time.
The comparison with the naive nearest-neighbor-based method (without attributes' importance) shows the effectiveness of the attributes' importance.
The performance of the baseline that uses random attributes' importance shows that the attributes' importance should be carefully designed.
In other words, our method enables users to incorporate appropriate domain knowledge on the target classes so that the performance of the model is enhanced.
Compared to DAP~\cite{lampert2014attribute}, the proposed method showed more stable performance on different poses.

The improvement compared to the best baseline (nearest neighbor without attributes' importance) was 4.55 points, which corresponds to 5.91\% relative improvement.
In addition, the proposed method achieved higher scores in majority of the poses compared to this baseline.
Especially a big improvement was observed in ``Stretching calf(L)'' and ``Stretching calf(R)'' poses.
This was because there were unignorable number of subjects who faced down when performing these poses though they were supposed to face forward according to the definition of the pose given in Table~\ref{tab:def}.
Our method could successfully deal with this intra-class variation simply by ignoring the status of head and focusing more on the other important attributes.

On the other hand, there are some poses whose F-measure dropped by incorporating the attributes' importance.
Among those, the biggest drop was observed in pose ``Folding arm''.
This was caused by the low estimation accuracy of the important attributes for folding-arm pose;
the statuses of shoulders in folding-arm pose were sometimes estimated as ``front'' while they had to be ``down'', probably~because arms were slightly pulled forward to make the space for hands at underarms.
Incorporating attributes' importance means focusing more on the important attributes for each pose and ignoring the other attributes.
Therefore, in case the attribute estimation accuracy is not good for those important attributes, the pose classification is done by relying too much on the unreliable attributes.
This problem may be addressed by integrating attributes' unreliability that was introduced~in \cite{jayaraman2014zero}.
\begin{table}[H]
\centering
\caption{The evaluation result (F-measure) of ZSL. Abbreviations are as follows. DAP: direct attribute prediction, NN: nearest neighbor, AI: attributes' importance. The bold numbers represent the best score or the one close to the best (the difference is less than 0.01) for each pose. }
\label{tab:reszsl}
\begin{tabular}{lcccc}\toprule
\textbf{Pose} & \textbf{DAP \cite{lampert2014attribute}} & \textbf{NN w/o AI}      & \textbf{NN w/random AI}               & \textbf{NN w/AI (Proposed)} \\
\midrule
Standing & {\bf 0.7148}  & 0.6953  & 0.3639  & 0.6944  \\
Sitting & 0.4072  & 0.6796  & 0.4567  & {\bf 0.7438}  \\
Squatting & 0.8922  & 0.9745  & 0.7637  & {\bf 1.0000}  \\
RaiseArmL & {\bf 1.0000}  & 0.9791  & 0.4212  & 0.9854  \\
RaiseArmR & {\bf 0.9973}  & 0.9662  & 0.4250  & 0.9522  \\
PointingL & {\bf 0.9937}  & 0.9541  & 0.4090  & 0.9721  \\
PointingR & 0.9629  & {\bf 0.9991}  & 0.4688  & {\bf 0.9947}  \\
FoldingArm & 0.4773  & {\bf 0.5345}  & 0.2206  & 0.4387  \\
DeepBreathing & 0.9061  & 0.9734  & 0.4457  & {\bf 0.9804}  \\
StretchingUp & {\bf 0.9913}  & 0.9861  & 0.5947  & {\bf 1.0000}  \\
StretchingForward & 0.3703  & {\bf 0.8778}  & 0.3625  & {\bf 0.8707}  \\
WaistBending & {\bf 1.0000}  & 0.9735  & 0.3795  & 0.9612  \\
WaistTwistingL & {\bf 0.4241}  & 0.2171  & 0.0995  & 0.2642  \\
WaistTwistingR & {\bf 0.3639}  & 0.1547  & 0.1150  & 0.2928  \\
HeelToBackL & {\bf 1.0000}  & {\bf 1.0000}  & 0.5458  & 0.9787  \\
HeelToBackR & {\bf 0.9931}  & 0.8710  & 0.4042  & 0.9729  \\
StretchingCalfL & 0.6647  & 0.5463  & 0.4160  & {\bf 0.8068}  \\
StretchingCalfR & {\bf 0.9570}  & 0.5956  & 0.4498  & 0.8897  \\
Boxing & 0.6549  & 0.6748  & 0.5434  & {\bf 0.7494}  \\
BaseballHitting & 0.5856  & 0.6957  & 0.4328  & {\bf 0.7739}  \\
Skiing & 0.5277  & {\bf 0.7977}  & 0.6334  & 0.7830  \\
Thinking & {\bf 0.8241}  & 0.7801  & 0.6420  & {\bf 0.8230}  \\ \midrule
avg. & 0.7595  & 0.7694  & 0.4361  & {\bf 0.8149}  \\ \bottomrule
\end{tabular}
\end{table}

\begin{figure}[H]
\begin{center}
\vspace{10pt}
\includegraphics[width=0.7\linewidth]{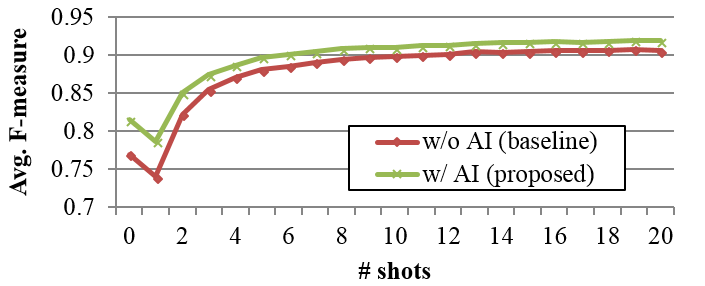}
\vspace{-20pt}
\caption{The results of few-shot learning. }
\label{fig:fsl}
\end{center}
\end{figure}

The result of few-shot learning is shown in Figure~\ref{fig:fsl}.
Here we only compared the proposed method with the best performed baseline, which is a nearest-neighbor-based method without attribute importance.
It shows that incorporating attributes' importance consistently improves the performance also in few-shot learning scenario.
The improvement is especially bigger when number of shots (training data) is less, and the impact of attributes' importance becomes smaller as number of available training data increases.
This is because the intra-class variation is reflected more in the training data as the number of available training increases and the classifier can naturally learn which attribute is actually important.
Another interesting observation is that the F-measure in ZSL scenario was better than that in one-shot learning scenario regardless of with or without attributes' importance.
This implies that under a situation where only extremely limited number of training data is available, human knowledge (pose definition table) can give a better compromise than just relying on the small number of training data.

\section{Conclusions}
\label{sec:conclusion}
This paper has presented a simple yet effective method for improving the performance of ZSL.
In~contrast to the conventional ZSL methods, the proposed method takes the importance of each attribute for each class into account, which becomes more critical when using a set of fine-grained attributes in order to represent wide variety of human poses and actions.
The experimental results on our dataset HDPoseDS have shown that the proposed method is effective not only for ZSL scenario, but also for few-shot learning scenario.
The results as well as the provided dataset are expected to promote further researches toward practical development of human-action-recognition technology under the situation of limited training data.

\vspace{6pt} 



\authorcontributions{
Conceptualization, H.O., M.A.-N., S.A., K.N., T.S., and A.D.;
Methodology, H.O., K.N., and T.S.;
Software, H.O.;
Formal Analysis, H.O.;
Investigation, H.O., M.A.-N., and S.A.;
Data Curation, H.O.;
Writing---Original Draft Preparation, H.O.;
Writing---Review \& Editing, H.O., S.A., and K.N.;
Supervision, A.D.
}

\funding{This research received no external funding.}


\conflictsofinterest{The authors declare no conflict of interest.}

%

\appendixtitles{yes} 
\appendixsections{multiple} 

\appendix
\section{Detailed Evaluation Results}
\label{sec:appendix}
{We show the }confusion matrices of the 4 methods mentioned in Section~\ref{sec:experiment}.
\tabcolsep = 2pt
\begin{table}[H]
\centering
\caption{The confusion matrix of the DAP~\cite{lampert2014attribute}.  The numbers in the first row and the first column correspond to the pose IDs shown in Table~\ref{tab:val}. T denotes total number in rows and columns. P and R denote precision and recall, respectively. The number in the bottom right is the accuracy (sum of diagonal elements divided by the total numbers). Please note that this is different from the F-measure.}
\label{tab:conf1}
\scalebox{0.72}[0.72]{
\begin{tabular}{lcccccccccccccccccccccccc}\toprule
& 1 & 2 & 3 & 4 & 5 & 6 & 7 & 8 & 9 & 10 & 11 & 12 & 13 & 14 & 15 & 16 & 17 & 18 & 19 & 20 & 21 & 22 & T & R \\ \midrule
1 & 584 & 0 & 0 & 0 & 0 & 0 & 0 & 0 & 0 & 0 & 0 & 0 & 0 & 0 & 0 & 0 & 0 & 0 & 0 & 0 & 0 & 0 & 584  & 1.00  \\ \midrule
2 & 48 & 135 & 126 & 0 & 0 & 0 & 0 & 71 & 13 & 0 & 0 & 0 & 0 & 0 & 0 & 0 & 0 & 0 & 0 & 19 & 6 & 110 & 528  & 0.26  \\ \midrule
3 & 0 & 0 & 538 & 0 & 0 & 0 & 0 & 0 & 0 & 0 & 0 & 0 & 0 & 0 & 0 & 0 & 0 & 0 & 0 & 0 & 1 & 3 & 542  & 0.99  \\ \midrule
4 & 0 & 0 & 0 & 540 & 0 & 0 & 0 & 0 & 0 & 0 & 0 & 0 & 0 & 0 & 0 & 0 & 0 & 0 & 0 & 0 & 0 & 0 & 540  & 1.00  \\ \midrule
5 & 0 & 0 & 0 & 0 & 548 & 0 & 0 & 0 & 0 & 0 & 0 & 0 & 0 & 0 & 0 & 0 & 0 & 0 & 0 & 0 & 0 & 0 & 548  & 1.00  \\ \midrule
6 & 0 & 0 & 0 & 0 & 0 & 553 & 0 & 0 & 0 & 0 & 0 & 0 & 0 & 0 & 0 & 0 & 0 & 0 & 0 & 0 & 0 & 0 & 553  & 1.00  \\ \midrule
7 & 0 & 0 & 0 & 0 & 0 & 0 & 571 & 0 & 0 & 0 & 0 & 0 & 0 & 0 & 0 & 0 & 0 & 0 & 0 & 0 & 0 & 0 & 571  & 1.00  \\ \midrule
8 & 0 & 0 & 0 & 0 & 0 & 0 & 0 & 257 & 0 & 0 & 0 & 0 & 0 & 0 & 0 & 0 & 154 & 0 & 169 & 0 & 0 & 0 & 580  & 0.44  \\ \midrule
9 & 0 & 0 & 0 & 0 & 0 & 0 & 0 & 0 & 531 & 0 & 0 & 0 & 0 & 0 & 0 & 0 & 0 & 0 & 0 & 0 & 0 & 0 & 531  & 1.00  \\ \midrule
10 & 0 & 0 & 0 & 0 & 0 & 0 & 0 & 0 & 0 & 569 & 0 & 0 & 0 & 0 & 0 & 0 & 0 & 0 & 0 & 0 & 0 & 0 & 569  & 1.00  \\ \midrule
11 & 0 & 0 & 0 & 0 & 0 & 0 & 0 & 0 & 0 & 10 & 127 & 0 & 0 & 0 & 0 & 0 & 419 & 0 & 0 & 0 & 0 & 0 & 556  & 0.23  \\ \midrule
12 & 0 & 0 & 0 & 0 & 0 & 0 & 0 & 0 & 0 & 0 & 0 & 522 & 0 & 0 & 0 & 0 & 0 & 0 & 0 & 0 & 0 & 0 & 522  & 1.00  \\ \midrule
13 & 197 & 0 & 0 & 0 & 3 & 0 & 44 & 77 & 11 & 0 & 3 & 0 & 148 & 0 & 0 & 0 & 0 & 0 & 27 & 0 & 31 & 0 & 541  & 0.27  \\ \midrule
14 & 221 & 0 & 0 & 0 & 0 & 0 & 0 & 88 & 86 & 0 & 0 & 0 & 9 & 121 & 0 & 0 & 0 & 0 & 19 & 0 & 0 & 0 & 544  & 0.22  \\ \midrule
15 & 0 & 0 & 0 & 0 & 0 & 0 & 0 & 0 & 0 & 0 & 0 & 0 & 0 & 0 & 527 & 0 & 0 & 0 & 0 & 0 & 0 & 0 & 527  & 1.00  \\ \midrule
16 & 0 & 0 & 0 & 0 & 0 & 7 & 0 & 0 & 0 & 0 & 0 & 0 & 0 & 0 & 0 & 505 & 0 & 0 & 0 & 0 & 0 & 0 & 512  & 0.99  \\ \midrule
17 & 0 & 0 & 0 & 0 & 0 & 0 & 0 & 0 & 0 & 0 & 0 & 0 & 0 & 0 & 0 & 0 & 568 & 0 & 0 & 0 & 0 & 0 & 568  & 1.00  \\ \midrule
18 & 0 & 0 & 0 & 0 & 0 & 0 & 0 & 0 & 0 & 0 & 0 & 0 & 0 & 0 & 0 & 0 & 0 & 568 & 0 & 0 & 0 & 0 & 568  & 1.00  \\ \midrule
19 & 0 & 0 & 0 & 0 & 0 & 0 & 0 & 0 & 0 & 0 & 0 & 0 & 0 & 0 & 0 & 0 & 0 & 0 & 539 & 0 & 0 & 0 & 539  & 1.00  \\ \midrule
20 & 0 & 0 & 0 & 0 & 0 & 0 & 0 & 0 & 0 & 0 & 0 & 0 & 0 & 0 & 0 & 0 & 0 & 51 & 110 & 219 & 127 & 0 & 507  & 0.43  \\ \midrule
21 & 0 & 0 & 0 & 0 & 0 & 0 & 0 & 4 & 0 & 0 & 0 & 0 & 0 & 0 & 0 & 0 & 0 & 0 & 243 & 3 & 291 & 0 & 541  & 0.54  \\ \midrule
22 & 0 & 0 & 0 & 0 & 0 & 0 & 0 & 0 & 0 & 0 & 0 & 0 & 0 & 0 & 0 & 0 & 0 & 0 & 0 & 0 & 106 & 513 & 619  & 0.83  \\ \midrule
T & 1050  & 135  & 664  & 540  & 551  & 560  & 615  & 497  & 641  & 579  & 130  & 522  & 157  & 121  & 527  & 505  & 1141  & 619  & 1107  & 241  & 562  & 626  & 12,090  &  \\ \midrule
P & 0.56  & 1.00  & 0.81  & 1.00  & 0.99  & 0.99  & 0.93  & 0.52  & 0.83  & 0.98  & 0.98  & 1.00  & 0.94  & 1.00  & 1.00  & 1.00  & 0.50  & 0.92  & 0.49  & 0.91  & 0.52  & 0.82  &  & 0.78  \\ \bottomrule
\end{tabular}
} 
\end{table}

\vspace{-6pt}

\begin{table}[H]
\centering
\caption{The confusion matrix of the nearest-neighbor-based baseline without attributes' importance.}
\label{tab:conf2}
\scalebox{0.72}[0.72]{
\begin{tabular}{lcccccccccccccccccccccccc}\toprule
& 1 & 2 & 3 & 4 & 5 & 6 & 7 & 8 & 9 & 10 & 11 & 12 & 13 & 14 & 15 & 16 & 17 & 18 & 19 & 20 & 21 & 22 & T & R \\ \midrule
1 & 543 & 0 & 0 & 0 & 0 & 0 & 0 & 0 & 0 & 0 & 0 & 0 & 41 & 0 & 0 & 0 & 0 & 0 & 0 & 0 & 0 & 0 & 584  & 0.93  \\ \midrule
2 & 0 & 316 & 22 & 0 & 0 & 0 & 0 & 57 & 0 & 0 & 0 & 0 & 0 & 0 & 0 & 0 & 0 & 0 & 0 & 1 & 72 & 60 & 528  & 0.60  \\ \midrule
3 & 0 & 1 & 536 & 0 & 0 & 0 & 0 & 0 & 0 & 0 & 0 & 0 & 0 & 0 & 0 & 0 & 0 & 0 & 0 & 0 & 0 & 5 & 542  & 0.99  \\ \midrule
4 & 0 & 0 & 0 & 540 & 0 & 0 & 0 & 0 & 0 & 0 & 0 & 0 & 0 & 0 & 0 & 0 & 0 & 0 & 0 & 0 & 0 & 0 & 540  & 1.00  \\ \midrule
5 & 9 & 0 & 0 & 0 & 515 & 0 & 0 & 0 & 0 & 8 & 11 & 0 & 5 & 0 & 0 & 0 & 0 & 0 & 0 & 0 & 0 & 0 & 548  & 0.94  \\ \midrule
6 & 0 & 0 & 0 & 2 & 0 & 551 & 0 & 0 & 0 & 0 & 0 & 0 & 0 & 0 & 0 & 0 & 0 & 0 & 0 & 0 & 0 & 0 & 553  & 1.00  \\ \midrule
7 & 0 & 0 & 0 & 0 & 1 & 0 & 570 & 0 & 0 & 0 & 0 & 0 & 0 & 0 & 0 & 0 & 0 & 0 & 0 & 0 & 0 & 0 & 571  & 1.00  \\ \midrule
8 & 0 & 0 & 0 & 0 & 0 & 0 & 0 & 306 & 0 & 0 & 33 & 0 & 65 & 0 & 0 & 0 & 0 & 0 & 176 & 0 & 0 & 0 & 580  & 0.53  \\ \midrule
9 & 0 & 0 & 0 & 0 & 0 & 0 & 0 & 0 & 531 & 0 & 0 & 0 & 0 & 0 & 0 & 0 & 0 & 0 & 0 & 0 & 0 & 0 & 531  & 1.00  \\ \midrule
10 & 0 & 0 & 0 & 0 & 0 & 0 & 0 & 0 & 0 & 569 & 0 & 0 & 0 & 0 & 0 & 0 & 0 & 0 & 0 & 0 & 0 & 0 & 569  & 1.00  \\ \midrule
11 & 0 & 0 & 0 & 0 & 0 & 0 & 0 & 0 & 0 & 8 & 535 & 0 & 0 & 0 & 0 & 0 & 13 & 0 & 0 & 0 & 0 & 0 & 556  & 0.96  \\ \midrule
12 & 0 & 0 & 0 & 0 & 0 & 0 & 0 & 0 & 0 & 0 & 0 & 495 & 0 & 3 & 0 & 0 & 9 & 12 & 0 & 0 & 3 & 0 & 522  & 0.95  \\ \midrule
13 & 195 & 0 & 0 & 11 & 2 & 0 & 0 & 60 & 17 & 0 & 0 & 0 & 108 & 121 & 0 & 0 & 0 & 0 & 27 & 0 & 0 & 0 & 541  & 0.20  \\ \midrule
14 & 165 & 0 & 0 & 10 & 0 & 0 & 0 & 51 & 12 & 0 & 4 & 0 & 190 & 56 & 0 & 0 & 0 & 0 & 56 & 0 & 0 & 0 & 544  & 0.10  \\ \midrule
15 & 0 & 0 & 0 & 0 & 0 & 0 & 0 & 0 & 0 & 0 & 0 & 0 & 0 & 0 & 527 & 0 & 0 & 0 & 0 & 0 & 0 & 0 & 527  & 1.00  \\ \midrule
16 & 66 & 0 & 0 & 0 & 0 & 51 & 0 & 0 & 0 & 0 & 0 & 0 & 0 & 0 & 0 & 395 & 0 & 0 & 0 & 0 & 0 & 0 & 512  & 0.77  \\ \midrule
17 & 0 & 0 & 0 & 0 & 0 & 0 & 0 & 0 & 0 & 0 & 53 & 0 & 0 & 0 & 0 & 0 & 298 & 217 & 0 & 0 & 0 & 0 & 568  & 0.52  \\ \midrule
18 & 0 & 0 & 0 & 0 & 0 & 0 & 0 & 0 & 0 & 0 & 27 & 0 & 0 & 0 & 0 & 0 & 203 & 338 & 0 & 0 & 0 & 0 & 568  & 0.60  \\ \midrule
19 & 0 & 0 & 0 & 0 & 0 & 0 & 0 & 43 & 0 & 0 & 0 & 0 & 0 & 0 & 0 & 0 & 0 & 0 & 496 & 0 & 0 & 0 & 539  & 0.92  \\ \midrule
20 & 0 & 0 & 0 & 0 & 0 & 0 & 0 & 48 & 0 & 0 & 0 & 0 & 0 & 0 & 0 & 0 & 0 & 0 & 170 & 272 & 16 & 1 & 507  & 0.54  \\ \midrule
21 & 0 & 4 & 0 & 0 & 0 & 0 & 0 & 0 & 0 & 0 & 0 & 0 & 45 & 0 & 0 & 0 & 0 & 0 & 6 & 1 & 485 & 0 & 541  & 0.90  \\ \midrule
22 & 0 & 81 & 0 & 0 & 0 & 0 & 0 & 0 & 0 & 0 & 0 & 0 & 0 & 0 & 0 & 0 & 0 & 0 & 0 & 1 & 99 & 438 & 619  & 0.71  \\ \midrule
T & 978  & 402  & 558  & 563  & 518  & 602  & 570  & 565  & 560  & 585  & 663  & 495  & 454  & 180  & 527  & 395  & 523  & 567  & 931  & 275  & 675  & 504  & 12,090  &  \\ \midrule
P & 0.56  & 0.79  & 0.96  & 0.96  & 0.99  & 0.92  & 1.00  & 0.54  & 0.95  & 0.97  & 0.81  & 1.00  & 0.24  & 0.31  & 1.00  & 1.00  & 0.57  & 0.60  & 0.53  & 0.99  & 0.72  & 0.87  &  & 0.78  \\ \bottomrule
\end{tabular}
} 
\end{table}

\begin{table}[H]
\centering
\caption{The confusion matrix of the nearest-neighbor-based baseline with random attributes'~importance. {The numbers} are the average of 1000 times. The precision and recall were calculated based on these averaged numbers.}
\label{tab:conf3}
\scalebox{0.73}[0.73]{
\begin{tabular}{lcccccccccccccccccccccccc}\toprule
& 1 & 2 & 3 & 4 & 5 & 6 & 7 & 8 & 9 & 10 & 11 & 12 & 13 & 14 & 15 & 16 & 17 & 18 & 19 & 20 & 21 & 22 & T & R \\ \midrule
1 & 331  & 0  & 0  & 42  & 49  & 30  & 40  & 6  & 14  & 7  & 5  & 13  & 12  & 4  & 12  & 11  & 2  & 1  & 5  & 0  & 0  & 0  & 584  & 0.57  \\ \midrule
2 & 17  & 206  & 78  & 31  & 8  & 13  & 4  & 28  & 3  & 1  & 1  & 5  & 5  & 4  & 7  & 4  & 1  & 4  & 8  & 8  & 42  & 48  & 528  & 0.39  \\ \midrule
3 & 8  & 35  & 437  & 6  & 2  & 2  & 2  & 3  & 1  & 0  & 0  & 1  & 1  & 0  & 8  & 5  & 0  & 0  & 1  & 1  & 5  & 24  & 542  & 0.81  \\ \midrule
4 & 56  & 0  & 1  & 303  & 24  & 41  & 17  & 9  & 15  & 27  & 11  & 4  & 8  & 3  & 3  & 8  & 3  & 2  & 4  & 0  & 0  & 0  & 540  & 0.56  \\ \midrule
5 & 66  & 0  & 0  & 26  & 293  & 19  & 39  & 5  & 11  & 25  & 21  & 14  & 7  & 3  & 8  & 2  & 5  & 3  & 1  & 0  & 0  & 0  & 548  & 0.53  \\ \midrule
6 & 69  & 1  & 1  & 77  & 26  & 245  & 18  & 16  & 22  & 15  & 14  & 5  & 11  & 6  & 8  & 8  & 2  & 3  & 7  & 1  & 0  & 0  & 553  & 0.44  \\ \midrule
7 & 69  & 0  & 0  & 25  & 57  & 21  & 291  & 9  & 23  & 6  & 20  & 10  & 12  & 5  & 6  & 7  & 5  & 3  & 3  & 1  & 0  & 0  & 571  & 0.51  \\ \midrule
8 & 42  & 2  & 2  & 46  & 28  & 33  & 20  & 109  & 19  & 13  & 43  & 4  & 38  & 19  & 7  & 8  & 13  & 7  & 103  & 11  & 13  & 1  & 580  & 0.19  \\ \midrule
9 & 45  & 0  & 0  & 50  & 45  & 32  & 36  & 12  & 225  & 12  & 16  & 9  & 16  & 8  & 5  & 6  & 5  & 4  & 3  & 2  & 0  & 0  & 531  & 0.42  \\ \midrule
10 & 14  & 0  & 0  & 46  & 29  & 13  & 7  & 3  & 5  & 354  & 53  & 4  & 1  & 1  & 5  & 2  & 18  & 11  & 0  & 0  & 0  & 0  & 569  & 0.62  \\ \midrule
11 & 21  & 0  & 0  & 23  & 35  & 17  & 28  & 7  & 9  & 54  & 233  & 6  & 3  & 1  & 3  & 2  & 74  & 40  & 0  & 1  & 0  & 0  & 556  & 0.42  \\ \midrule
12 & 78  & 1  & 0  & 27  & 55  & 16  & 29  & 4  & 14  & 9  & 12  & 168  & 20  & 12  & 25  & 10  & 20  & 14  & 3  & 1  & 3  & 0  & 522  & 0.32  \\ \midrule
13 & 118  & 1  & 1  & 44  & 44  & 30  & 36  & 37  & 32  & 8  & 14  & 26  & 41  & 43  & 12  & 15  & 4  & 4  & 24  & 4  & 4  & 0  & 541  & 0.08  \\ \midrule
14 & 83  & 0  & 1  & 38  & 43  & 36  & 31  & 41  & 33  & 11  & 17  & 26  & 62  & 42  & 15  & 16  & 7  & 4  & 26  & 8  & 4  & 1  & 544  & 0.08  \\ \midrule
15 & 72  & 2  & 6  & 32  & 28  & 24  & 13  & 5  & 9  & 11  & 6  & 23  & 8  & 3  & 263  & 6  & 3  & 4  & 6  & 3  & 1  & 0  & 527  & 0.50  \\ \midrule
16 & 92  & 2  & 7  & 40  & 29  & 37  & 26  & 9  & 27  & 9  & 11  & 17  & 11  & 3  & 11  & 162  & 8  & 4  & 6  & 0  & 1  & 0  & 512  & 0.32  \\ \midrule
17 & 8  & 0  & 0  & 7  & 11  & 5  & 9  & 1  & 4  & 27  & 134  & 8  & 1  & 1  & 2  & 2  & 232  & 116  & 0  & 0  & 0  & 0  & 568  & 0.41  \\ \midrule
18 & 6  & 0  & 0  & 8  & 9  & 4  & 6  & 2  & 5  & 26  & 109  & 7  & 2  & 1  & 2  & 1  & 143  & 234  & 0  & 1  & 0  & 0  & 568  & 0.41  \\ \midrule
19 & 16  & 1  & 1  & 11  & 5  & 13  & 8  & 51  & 6  & 2  & 1  & 3  & 12  & 7  & 5  & 3  & 0  & 0  & 360  & 17  & 17  & 1  & 539  & 0.67  \\ \midrule
20 & 7  & 5  & 3  & 7  & 3  & 6  & 4  & 34  & 3  & 4  & 5  & 3  & 7  & 8  & 21  & 1  & 2  & 10  & 138  & 165  & 57  & 15  & 507  & 0.32  \\ \midrule
21 & 10  & 29  & 4  & 6  & 4  & 5  & 4  & 12  & 1  & 1  & 1  & 7  & 9  & 4  & 7  & 4  & 1  & 0  & 64  & 15  & 343  & 9  & 541  & 0.63  \\ \midrule
22 & 7  & 87  & 61  & 5  & 2  & 2  & 4  & 6  & 1  & 0  & 1  & 2  & 2  & 1  & 3  & 5  & 0  & 2  & 24  & 16  & 50  & 340  & 619  & 0.55  \\ \midrule
T & 1236  & 373  & 603  & 900  & 829  & 643  & 669  & 407  & 480  & 622  & 728  & 365  & 290  & 178  & 437  & 288  & 548  & 473  & 787  & 254  & 541  & 439  & 12,090  &  \\ \midrule
P & 0.27  & 0.55  & 0.72  & 0.34  & 0.35  & 0.38  & 0.43  & 0.27  & 0.47  & 0.57  & 0.32  & 0.46  & 0.14  & 0.23  & 0.60  & 0.56  & 0.42  & 0.50  & 0.46  & 0.65  & 0.63  & 0.77  &  & 0.44  \\ \bottomrule
\end{tabular}
} 
\end{table}

\vspace{-6pt}

\begin{table}[H]
\centering
\caption{The confusion matrix of the proposed method.}
\label{tab:conf4}
\scalebox{0.73}[0.73]{
\begin{tabular}{lcccccccccccccccccccccccc}\toprule
& 1 & 2 & 3 & 4 & 5 & 6 & 7 & 8 & 9 & 10 & 11 & 12 & 13 & 14 & 15 & 16 & 17 & 18 & 19 & 20 & 21 & 22 & T & R \\ \midrule
1 & 526 & 0 & 0 & 0 & 0 & 0 & 0 & 0 & 0 & 0 & 0 & 0 & 58 & 0 & 0 & 0 & 0 & 0 & 0 & 0 & 0 & 0 & 584  & 0.90  \\ \midrule
2 & 8 & 421 & 0 & 0 & 0 & 0 & 0 & 46 & 0 & 0 & 0 & 0 & 0 & 0 & 0 & 0 & 0 & 1 & 0 & 4 & 12 & 36 & 528  & 0.80  \\ \midrule
3 & 0 & 0 & 542 & 0 & 0 & 0 & 0 & 0 & 0 & 0 & 0 & 0 & 0 & 0 & 0 & 0 & 0 & 0 & 0 & 0 & 0 & 0 & 542  & 1.00  \\ \midrule
4 & 0 & 0 & 0 & 540 & 0 & 0 & 0 & 0 & 0 & 0 & 0 & 0 & 0 & 0 & 0 & 0 & 0 & 0 & 0 & 0 & 0 & 0 & 540  & 1.00  \\ \midrule
5 & 16 & 0 & 0 & 0 & 498 & 0 & 0 & 0 & 0 & 0 & 0 & 0 & 34 & 0 & 0 & 0 & 0 & 0 & 0 & 0 & 0 & 0 & 548  & 0.91  \\ \midrule
6 & 0 & 0 & 0 & 1 & 0 & 523 & 0 & 0 & 0 & 0 & 0 & 0 & 28 & 1 & 0 & 0 & 0 & 0 & 0 & 0 & 0 & 0 & 553  & 0.95  \\ \midrule
7 & 0 & 0 & 0 & 0 & 0 & 0 & 565 & 0 & 3 & 0 & 0 & 0 & 3 & 0 & 0 & 0 & 0 & 0 & 0 & 0 & 0 & 0 & 571  & 0.99  \\ \midrule
8 & 0 & 0 & 0 & 5 & 0 & 0 & 0 & 188 & 0 & 0 & 0 & 0 & 179 & 120 & 0 & 0 & 0 & 0 & 88 & 0 & 0 & 0 & 580  & 0.32  \\ \midrule
9 & 0 & 0 & 0 & 0 & 0 & 0 & 0 & 0 & 525 & 0 & 0 & 0 & 6 & 0 & 0 & 0 & 0 & 0 & 0 & 0 & 0 & 0 & 531  & 0.99  \\ \midrule
10 & 0 & 0 & 0 & 0 & 0 & 0 & 0 & 0 & 0 & 569 & 0 & 0 & 0 & 0 & 0 & 0 & 0 & 0 & 0 & 0 & 0 & 0 & 569  & 1.00  \\ \midrule
11 & 0 & 0 & 0 & 0 & 0 & 0 & 0 & 0 & 0 & 0 & 478 & 0 & 2 & 0 & 0 & 0 & 72 & 4 & 0 & 0 & 0 & 0 & 556  & 0.86  \\ \midrule
12 & 36 & 0 & 0 & 0 & 0 & 0 & 0 & 0 & 0 & 0 & 0 & 483 & 0 & 0 & 0 & 0 & 0 & 0 & 0 & 0 & 3 & 0 & 522  & 0.93  \\ \midrule
13 & 193 & 0 & 0 & 10 & 0 & 0 & 0 & 10 & 9 & 0 & 0 & 0 & 191 & 191 & 0 & 0 & 0 & 0 & 1 & 0 & 0 & 0 & 605  & 0.32  \\ \midrule
14 & 103 & 0 & 0 & 0 & 0 & 0 & 0 & 5 & 3 & 0 & 6 & 0 & 291 & 136 & 0 & 0 & 0 & 0 & 0 & 0 & 0 & 0 & 544  & 0.25  \\ \midrule
15 & 22 & 0 & 0 & 0 & 0 & 0 & 0 & 0 & 0 & 0 & 0 & 0 & 0 & 0 & 505 & 0 & 0 & 0 & 0 & 0 & 0 & 0 & 527  & 0.96  \\ \midrule
16 & 27 & 0 & 0 & 0 & 0 & 0 & 0 & 0 & 0 & 0 & 0 & 0 & 0 & 0 & 0 & 485 & 0 & 0 & 0 & 0 & 0 & 0 & 512  & 0.95  \\ \midrule
17 & 0 & 0 & 0 & 0 & 0 & 0 & 0 & 0 & 0 & 0 & 51 & 0 & 2 & 0 & 0 & 0 & 449 & 66 & 0 & 0 & 0 & 0 & 568  & 0.79  \\ \midrule
18 & 0 & 0 & 0 & 0 & 0 & 0 & 0 & 0 & 0 & 0 & 7 & 0 & 24 & 1 & 0 & 0 & 24 & 512 & 0 & 0 & 0 & 0 & 568  & 0.90  \\ \midrule
19 & 0 & 0 & 0 & 0 & 0 & 0 & 0 & 28 & 0 & 0 & 0 & 0 & 37 & 0 & 0 & 0 & 0 & 0 & 474 & 0 & 0 & 0 & 539  & 0.88  \\ \midrule
20 & 0 & 0 & 0 & 0 & 0 & 0 & 0 & 0 & 0 & 0 & 0 & 0 & 0 & 0 & 0 & 0 & 0 & 0 & 163 & 344 & 0 & 0 & 507  & 0.68  \\ \midrule
21 & 0 & 69 & 0 & 0 & 0 & 0 & 0 & 0 & 0 & 0 & 0 & 0 & 50 & 0 & 0 & 0 & 0 & 0 & 0 & 34 & 388 & 0 & 541  & 0.72  \\ \midrule
22 & 0 & 114 & 0 & 0 & 0 & 0 & 0 & 0 & 0 & 0 & 0 & 0 & 0 & 0 & 0 & 0 & 0 & 0 & 0 & 0 & 47 & 458 & 619  & 0.74  \\ \midrule
T & 931  & 604  & 542  & 556  & 498  & 523  & 565  & 277  & 540  & 569  & 542  & 483  & 905  & 449  & 505  & 485  & 545  & 583  & 726  & 382  & 450  & 494  & 12,154  &  \\ \midrule
P & 0.56  & 0.70  & 1.00  & 0.97  & 1.00  & 1.00  & 1.00  & 0.68  & 0.97  & 1.00  & 0.88  & 1.00  & 0.21  & 0.30  & 1.00  & 1.00  & 0.82  & 0.88  & 0.65  & 0.90  & 0.86  & 0.93  &  & 0.81  \\ \bottomrule
\end{tabular}
} 
\end{table}
\tabcolsep = 6pt

\reftitle{References}

\sampleavailability{The experimental data used in this study are available from the authors at \url{http://projects.dfki.uni-kl.de/zsl/data/}.}

\end{document}